# Characterizing Lidar Range-Measurement Ambiguity due to Multiple Returns

Jason H. Rife and Yifan Li, *Tufts University*

**BIOGRAPHY**

**Jason Rife** is a Professor and Chair of the Department of Mechanical Engineering at Tufts University in Medford, Massachusetts. He directs the Automated Systems and Robotics Laboratory (ASAR), which applies theory and experiment to characterize integrity of autonomous vehicle systems. He received his B.S in Mechanical and Aerospace Engineering from Cornell University and his M.S. and Ph.D. degrees in Mechanical Engineering from Stanford University.

**Yifan Li** is a student in the Electrical and Computer Engineering Ph.D. program at Tufts University in Medford, MA. He works in the Automated Systems and Robotics Laboratory (ASAR) with Dr. Jason Rife. He received his B.S. degree in Information Engineering from Southern University of Science and Technology (China) and M.S. degree in Electrical Engineering from Tufts University.

**ABSTRACT**

Reliable position and attitude sensing is critical for highly automated vehicles that operate on conventional roadways. Lidar sensors are increasingly incorporated into pose-estimation systems. Despite its great utility, lidar is a complex sensor, and its performance in roadway environments is not yet well understood. For instance, it is often assumed in lidar-localization algorithms that a lidar will always identify a unique surface along a given raypath. However, this assumption is not always true, as ample prior evidence exists to suggest that lidar units may generate measurements probabilistically when more than one scattering surface appears within the lidar's conical beam. In this paper, we analyze lidar datasets to characterize cases with probabilistic returns along particular raypaths. Our contribution is to present representative cumulative distribution functions (CDFs) for raypaths observed by two different mechanically rotating lidar units with stationary bases. In subsequent discussion, we outline a qualitative methodology to assess the effect of probabilistic multi-return cases on lidar-based localization.

## 1 INTRODUCTION

Lidar has significant potential as a sensor for safety-critical navigation. In autonomous driving, for example, lidar is valuable for sensing dynamic, complex scenes and providing reliable localization even in varying lighting or weather conditions. However, an open research challenge related to analyzing lidar-system safety involves characterization of measurement-error distributions. The main goal of this paper is to support error characterization by analyzing lidar ranging data.

In particular, this paper focuses on multi-peak measurement distributions occurring when the conical lidar beam illuminates multiple scatterers along the same raypath. Multiple returns are a special case of multipath, a well-known cause of errors in radio navigation, including in satellite navigation systems (Braasch, 1994; Braasch, 1996; Chen et al., 2013; Xu & Rife, 2020). Multipath occurs when a receiver observes more than one copy of the expected signal, such as in GPS, when one ranging signal arrives directly from a satellite and a second arrives indirectly after reflecting from a wall or the ground. For a lidar detector array, it is possible for a signal to be received from an angle offset from signal-propagation axis (Cao et al., 2023; Nemana et al., 2025); however, a more common form of lidar multipath involves *multiple returns*, which occurs when lidar emissions scatter from more than one surface inside the conical beam associated with the direct raypath. For example, multiple returns can be generated at the corner of a structure when a lidar beam strikes two surfaces, as shown in Figure 1.

For safety-critical localization, multiple returns are an adversity that introduces measurement uncertainty and threatens sensor integrity (Rife et al., 2024). For instance, in performing vehicle positioning using geometric scan matching algorithms like ICP (Besl & McKay, 1992), NDT (Biber & Straßer, 2003), LOAM (Zhang & Singh, 2014), or ICET (McDermott & Rife, 2024), returns from surfaces other than the expected surface can bias registration of the current lidar scan to a reference (e.g. a prior scan or a map). Without a better understanding of the probabilistic nature of multiple returns, it will be difficult to conduct a rigorous safety analysis for lidar-based positioning for roadways or other unstructured environments.

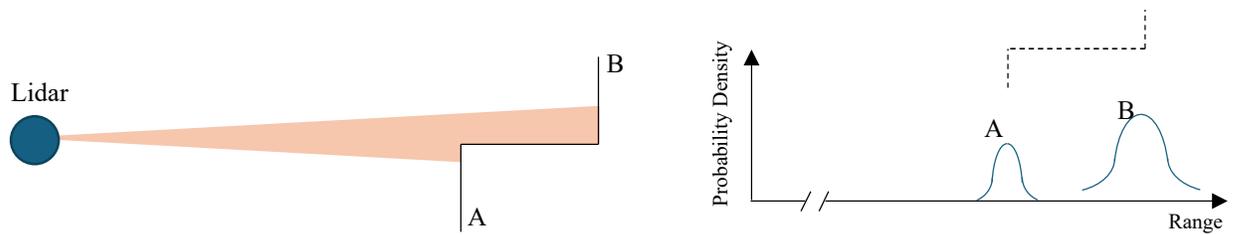

**FIGURE 1**

*Lidar beams have finite width. The left side of the figure shows a lidar beam intersecting two surfaces. Whenever the lidar samples a range measurement for this beam path, the lidar may report the range of the near surface (labeled A) or that of the far surface (labeled B). The right side of the figure sketches the probability density function (PDF) for the return measurement; probability density is shown as a function of the measured range. The left peak of the PDF corresponds to a range measurement from the near surface and the right peak, from the far surface.*

Though multiple returns are problematic for precise localization, they are a positive design feature in other applications. For instance, terrain mapping aerial lidar can be set to record only the final return, a setting that allows the lidar to peer through a sparse tree canopy and measure the ground elevation below. With a wide range of applications in mind, manufacturers sometimes allow users to customize lidar units to detect either the strongest return or the last return (Velodyne, 2019). Newer model lidars may record multiple returns (Ouster, 2022). For automotive lidar applications (Li & Ibanez-Guzman, 2020; Roriz et al., 2020; McDermott & Rife, 2022), the strongest-return option is generally preferred. This option increases the probability that a return scatters from a single dominant scatterer.

In this paper, we seek to understand better how instances of multiple returns impact the lidar measurement-error distribution. The paper's primary contribution is, for the first time, to compare cumulative distribution functions (CDFs) for different parts of scenes visualized by a mechanically rotating lidar. This type of lidar unit scans the same pattern of beam angles each time it spins, over a regular grid of azimuth and elevation angles. The collection of range values on this grid is sometimes called a *range image*, where each *pixel* provides a range measurement for a particular raypath. When viewing a series of range images for a stationary lidar, the range values for certain pixels may noticeably jitter, an indication of multiple returns. By compiling data for these jittering pixels, it is possible to assemble a CDF of the range measurement distribution for a given raypath. This paper uses data to characterize several cases where probabilistic measurements occur (e.g. for windows and foliage). Spatial CDFs are also considered, with the idea the spatial CDF over the neighborhood around a raypath may provide instantaneous clues about the nature of a temporal CDF.

The remainder of this paper is organized as follows. In the next section, we describe the datasets and analyses we used to generate meaningful CDFs describing probabilistic, multi-return cases. In our analysis, we focus on temporal CDFs for a particular raypath, where data is compiled over subsequent frames. For comparison, we also consider spatial CDFs drawn from the neighborhood of the range image surrounding the raypath of interest. A results section compares the CDFs for several representative examples of multi-return raypaths. A discussion section then notes the similarities between temporal and spatial CDFs, pointing the way to a potential detector for multi-return raypaths. We conclude the paper with a summary of key results.

## 2  METHODOLOGY

In this section, we describe our analysis techniques for processing lidar data to extract evidence of multiple returns along the same raypath. Specifically, we detail two datasets and the statistical methods we used to analyze construct empirical CDFs. We also describe a motion-compensation strategy to assist in cases when the lidar base moves slightly between frames.

### 2.1  Data

This paper considers two lidar sequences generated by two different models of mechanically spinning lidar. One sequence was drawn from the Newer College Dataset (Ramezani et al., 2020). The other sequence was drawn from a multimodal dataset compiled by the US Department of Transportation (USDOT)Volpe center (Wassaf et al., 2021).

First consider the Newer College Dataset, where we focus specifically on the Dynamic Spinning sequence (Table 1, entry 6). The Newer College Dataset was collected by an Ouster OS1 (Gen 1), a mechanically spinning lidar. Each rotation of the lidar, at 10 Hz, captures a single range image or *frame*. Frames contain 64 vertical (elevation) channels, each with 1024 horizontal (azimuth) samples. In the Dynamic Spinning data set, the OS1 visualizes an interior courtyard of the University of Oxford. The full sequence includes aggressive rotations (up to 3.5 rad/s); however, we consider only the opening sequence (first 30 frames), where the handheld lidar is held stationary by a research, who stands near one interior corner of the courtyard. For the stationary segment of the data, we examine four raypaths of interest: labeled *wall* (raypath strikes a flat wall), *corner* (raypath observes the right-angle between two flat walls), *window* (raypath extends through a glass window into a room interior), and *foliage* (raypath passes through a tree). The four cases are indicated in Figure 2, each with a red "x" marker.

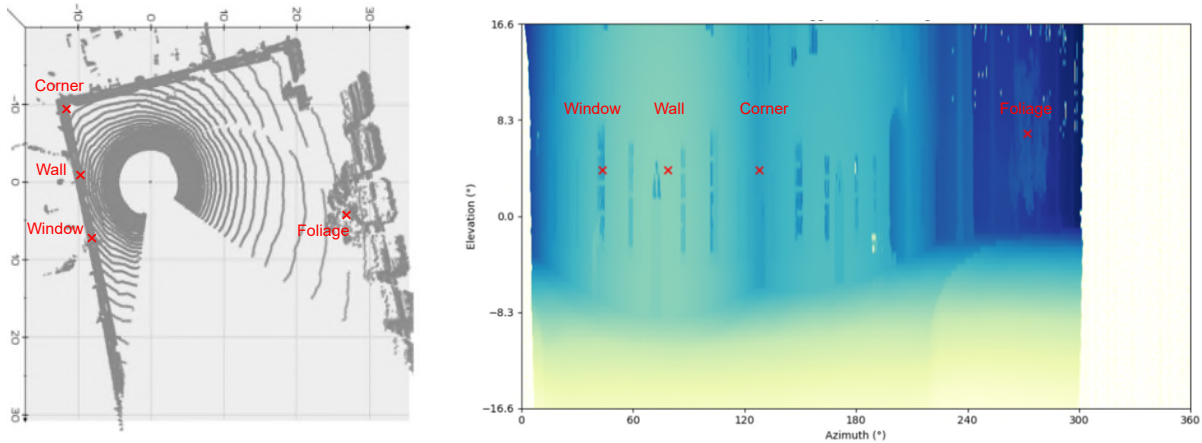

**FIGURE 2**

*Newer College Dynamic Spinning data: A point cloud (left) and range image (right) depict the interior of an academic courtyard. Four raypaths (red markers) are considered. In clockwise order these are: window, wall, corner, and foliage. The researcher holding the lidar has been removed by trimming data from an azimuthal arc of roughly 60 degrees.*

Figure 2 employs two modalities for visualizing lidar data: a point cloud and a range image. The 3D point cloud (left) shows the geometry of the physical scene. Here the point cloud has been rotated to show the bird's eye perspective, looking down on the courtyard from above. The range image visualization (right) emphasizes that the sample grid, with uniform sampling over a set of azimuth and elevation angles. data are sampled in a 2D fashion from a 3D world, with one sample acquired for each bin of azimuth and elevation. The range image is a distorted Mercator projection of the scene. In the range image, brighter colors (yellows) are closer and darker colors (blues) are farther away.

Second consider the USDOT Volpe dataset, where we focus specifically on the Signage sequence. This sequence was acquired in a parking lot in northern Virginia, USA in March of 2021. A Velodyne VLP-16 lidar unit captured the sequence. The VLP-16 was roof-mounted on a Mercedes van and configured to spin at 10 Hz. We consider a set of 50 frames early in the dataset when the van and lidar base are stationary. The VLP-16 has sixteen vertical (elevation) channels, each sampling 1800 horizontal (azimuth) bins. The parking lot scene from the sequence of interest is illustrated in Figure 3. Figure 3(a) is a range image, where the foreground pixels are bright and background pixels, dark. Non-returns are shaded black. The bright pixels in the lower half of the image are returns from the van roof. Figure 3(b) shows the same scene as a range image, viewed from above. The green square marker in the point cloud indicates the lidar unit, atop the data collection van. For comparison, Figure 3(c) shows the same scene viewed from above, via a satellite photograph. Finally, Figure 3(d) shows a photograph of the scene from a conventional camera located on the back of the van. This view is similar to the range image of Figure 3(a); however, because the camera does not provide a full 360° view around the van, the van has been displaced slightly forward to provide a broader view, corresponding to range-image azimuth angles from roughly -30° to 210° (where 90° points directly behind the van). Notably, the conventional camera image of Figure 3(d) reveals that tree leaf coverage is thin, a factor which is not otherwise obvious from the lidar data. Thin leaf coverage is expected for northern Virginia in March.

Six raypaths of interest were extracted from the Signage sequence, as indicated by red "x" markers in Figure 3(a). The raypaths are labeled 1 through 6. Raypath 1 corresponds to a hedgerow along the side of the parking lot. Raypaths 2 and 3 correspond to a large tree outside the hedgerow. Raypaths 4 through 6 correspond to features on an island in the middle of the parking lot. Specifically, raypaths 4 and 6 pass through two neighboring trees. Raypath 5 strikes a sign located between the two trees.

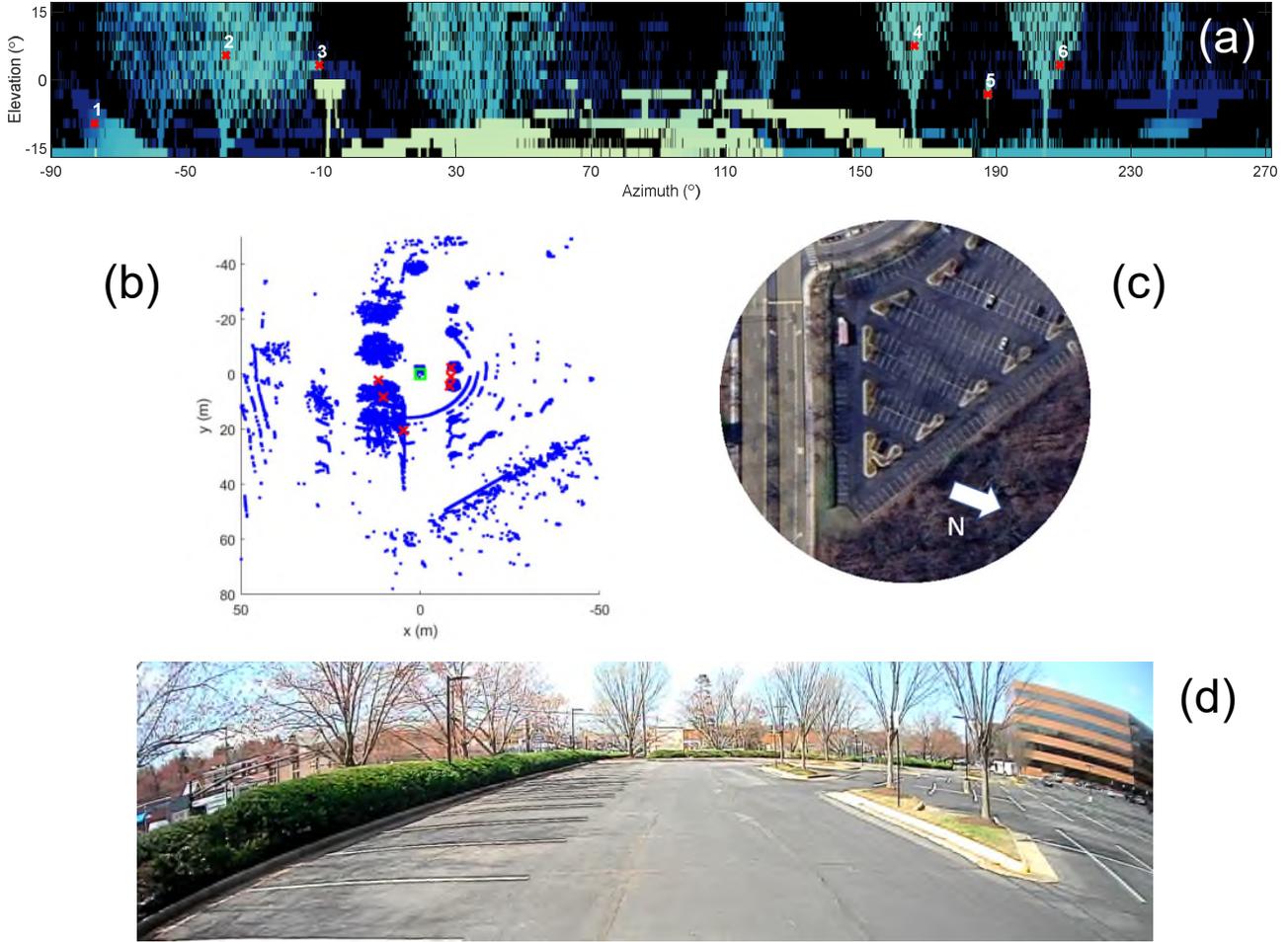

**FIGURE 3**
*Volpe Signage Dataset: The dataset was acquired in a parking lot. The range image (a) shows six enumerated raypaths of interest, each marked with a red x. The returns along these raypaths are also shown as red x markers within a 3D point cloud (b). Both the point cloud and an accompanying satellite photo (c) show the parking lot viewed from above. A photograph of the scene (d), taken from a rearward facing camera on the van, shows roughly the same perspective seen in the middle section of the lidar range image.*

## 2.2 Distributions

We inherently consider the distribution of range values along a given raypath to be probabilistic. As such, we collected multiple samples along each raypath over time (assuming a static sensor) to compile a list of range values. We then constructed an empirical cumulative distribution function (CDF) from the results. This CDF, labeled *F*, sums step functions for range measurements $d_{i,j,k}$ from each time step *k* sampled for a given pixel (i,j). For a set of K sequential time steps (starting at zero), this temporal CDF is

$$F(x; i, j) = \frac{1}{K} \sum_{k=0}^{K-1} \mathbb{1}(x - d_{i,j,k}) , \qquad (1)$$

where the CDF increases monotonically with increasing range *x* due to the accumulation of step functions. Step functions jump where the range *x* passes a measurement, where $x - d_{i,j,k} = 0$, given the step function definition is

$$\mathbb{1}(y) = \begin{cases} 0 & y < 0 \\ 1 & y \geq 0 \end{cases} . \qquad (2)$$

We also construct instantaneous spatial CDFs over neighborhood around a given raypath. In our analysis we define the neighborhood to consist of a 5x5 grid for the Newer college data and a 3x3 grid for the Volpe data (see Figure 4), with the center pixel associated with the raypath (i,j). For the 5x5 grid, the instantaneous spatial CDF around ray (i,j) at time $k$ is

$$G(x; i, j, k) = \frac{1}{25} \sum_{m=-2}^{2} \sum_{n=-2}^{2} \mathbb{1}(x - d_{i+m, j+n, k}). \qquad (3)$$

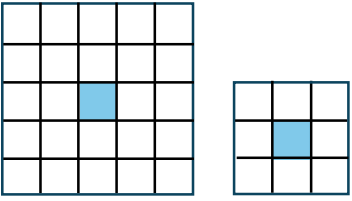

**FIGURE 4**
*Neighborhoods used to create spatial distributions: For the Newer College Dataset, which featured 64 vertical channels, we defined the neighborhood around the central raypath (shaded) using a 5x5 grid (left) spanning approximately 2° in elevation and 1.5° in azimuth. For the Volpe Dataset, which featured 16 vertical channels, we defined the neighborhood around the central raypath (shaded) using a 3x3 grid (right) spanning approximately 4° in elevation and 0.6° in azimuth.*

### 2.3 Motion Compensation

The temporal CDF, as defined in (1), assumes a single raypath can be sampled repeatedly. For a stationary lidar unit that produces a structured output, it is reasonable to sample a particular pixel (i,j) repeatedly as a way to obtain multiple readings for the same raypath. However, the lidar in the Dynamic Spinning Dataset is only approximately stationary. The lidar is held in the hands of a researcher, and small vibrations cause motion jitter. To assess whether this jitter impacts the temporal CDF, we implemented a motion compensation algorithm.

Specifically, we implemented patch-based motion-compensation for a 5x5 set of pixels (see Figure 4, left side). For each sequential frame $k$ patches centered at several points (p,q) are compared to patch from the original frame, at time 0, centered on the raypath (i,j). For a quality match, the measured range values for the two patches should be nearly the same, with a difference close to zero. Summing the squared range differences over the patch gives a cost metric $J$, for which a low value indicates an overall good match:

$$J(p, q; i, j, k) = \frac{1}{25} \sum_{m=-2}^{2} \sum_{n=-2}^{2} (d_{p+m, q+n, k} - d_{i+m, j+n, 0})^2. \qquad (4)$$

We considered (p,q) indices within two pixels of the original raypath index (i,j). Over the set of possible (p,q), the one minimizing the cost function at time $k$ is designated as the optimal match $(p_k^*, q_k^*)$:

$$(p_k^*, q_k^*) = \underset{\substack{p \in [i-2, i+2] \\ q \in [j-2, j+2]}}{\operatorname{argmin}} (J(p, q; i, j, k)). \qquad (5)$$

The motion-compensated CDF $\bar{F}$ is obtained by sampling the range value from the optimal-match pixel $(p_k^*, q_k^*)$ at each time step $k$. Note that at the original time step, $(p_k^*, q_k^*)$ is equal to (i,j).

$$\bar{F}(x; i, j) = \frac{1}{K} \sum_{k=0}^{K-1} \mathbb{1}(x - d_{p_k^*, q_k^*, k}), \qquad (6)$$

An example of this motion compensation process is shown in **Figure 5**.

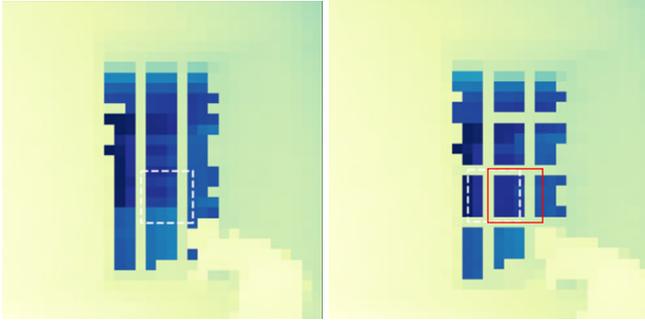

**FIGURE 5**
*Range-image visualization of a window at two times in the Newer College sequence. At time zero (left) the raypath of interest lies at the middle of the dashed square. At time k (right), the same pixel indices (dashed square) do not align well. Motion-compensation evaluates the optimal alignment requires a shift of two pixels to the right (red box).*

## 3 RESULTS

This section presents empirical CDFs for raypaths from two datasets. We focus first on temporal CDFs, computed with (1). Next, we compare spatial CDFs with (3) and compare them to temporal CDFs. A final set of results involves a comparison of CDFs generated with and without motion compensation.

### 3.1 Temporal CDFs

For the Newer College Dataset, we consider four raypaths. Temporal CDFs consist of ranges compiled from a 30 frame sequence for each raypath. The CDF for each raypath is shown in Figure 6. By construction, the empirical distributions increment from a minimum value of 1/30 to a maximum value of 1. Motion correction was not applied.

The top two CDFs in the figure (the *wall* and *corner* cases) both resemble step functions. The samples for the wall case are particularly well clustered; in quantitative terms, the wall-case range have a standard deviation of only 2 cm. By comparison, the samples for the interior corner are clustered nearly well, but with a heavy tail toward longer range values, where the heavy tail appears shifted by about 10 cm from the distribution core. The heavy tail may be caused in part by imprecision in azimuth sampling, with a few measurements acquired deeper into the corner than others. Another possible explanation for the heavy tail is that the raypath may reflect from one corner to the other, before scattering back to the detector. More data would be needed to differentiate between these possibilities.

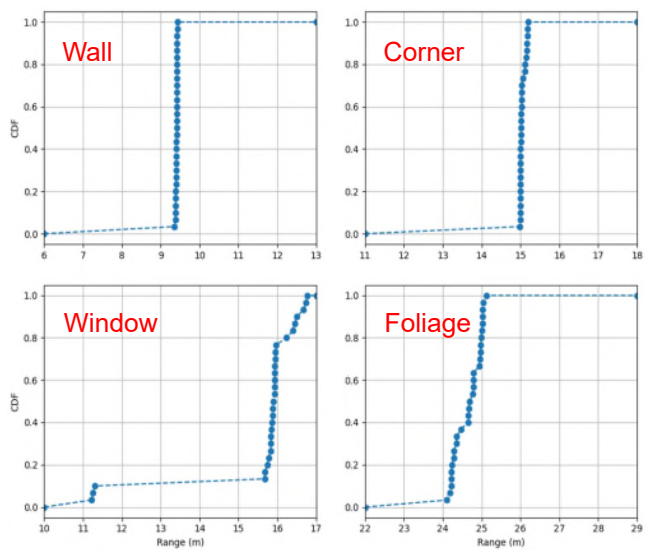

**FIGURE 6**
*Temporal CDFs obtained for four raypaths from the Newer College Dynamic Spinning sequence*

The bottom two CDFs in the figure (the *window* and *foliage* cases) are wider than the top two distributions. In the case of the window CDF, it appears that three returns (near 11.2 m in range) reflected from the window surface. The remaining returns have longer ranges, suggesting they were scattered by the room interior. These points also appear inside the building in the 3D point cloud (see outlier points near the marker labeled *window* in Figure 2). In the foliage case, the CDF also shows evidence of multiple scatterers, likely a mix of leaves and branches that spread range returns over the span of approximately one meter.

For the Volpe Dataset, we considered six raypaths. Temporal CDFs for each raypath are plotted in Figure 7. Motion correction was not applied in generating these CDFs. Because the lidar sequence consist of 50 frames, the empirical CDF increments from a minimum value of 1/50 to a maximum value of 1. Unlike the Newer College CDFs in Figure 6, however, not all of the CDFs in Figure 7 reach a maximum value of 1. The vertical space atop each CDF indicates the number of non-returns, which occur when no significant power spike is recorded before the detector times out.

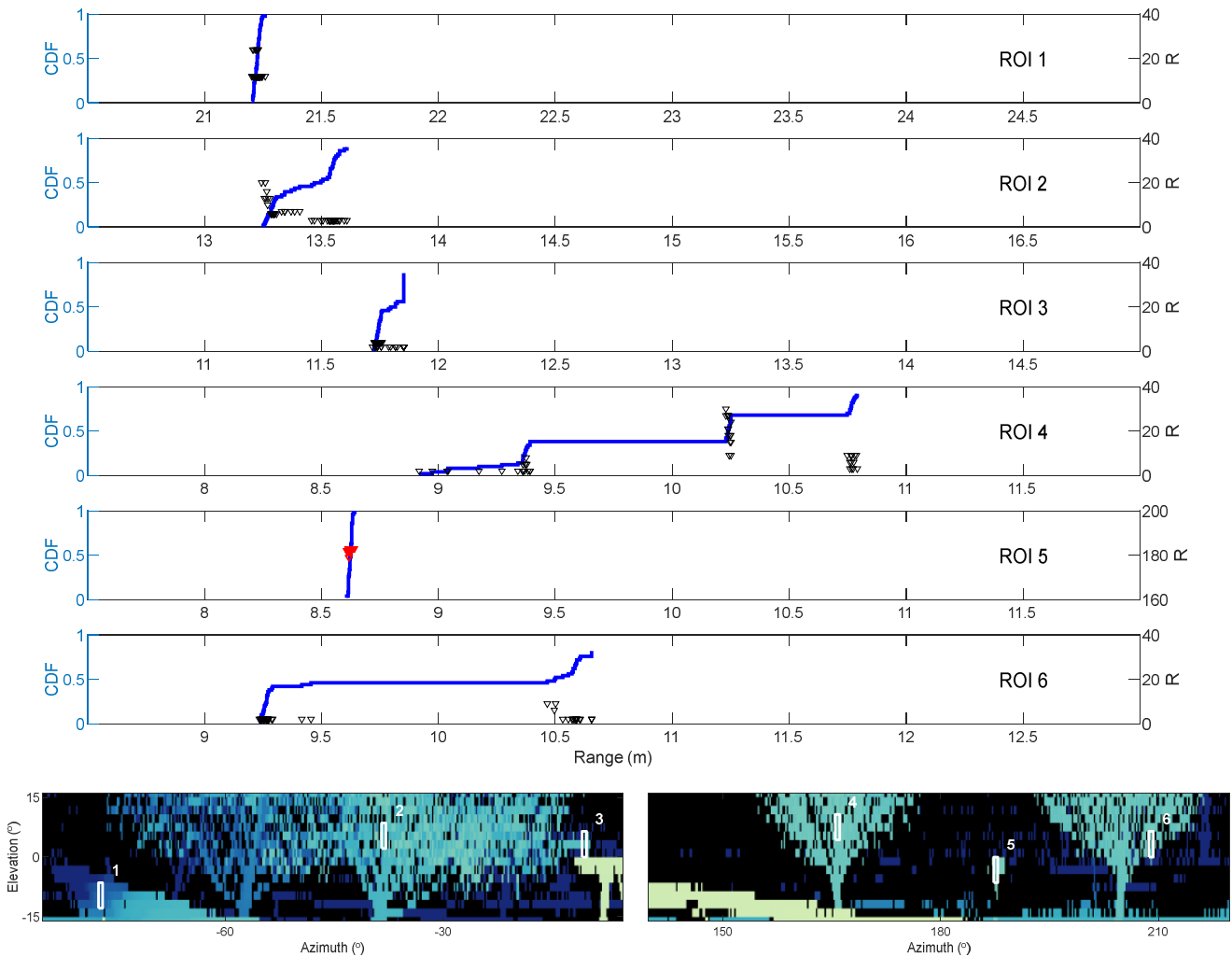

**FIGURE 7**
*Temporal CDFs obtained for six raypaths from the Volpe Signage sequence. The raypaths are labeled as Regions of Interest (ROI) 1 through 6. A CDFs (blue) is shown for each raypath. A plot of reflectance (right vertical axis) versus range (horizontal axis) is superposed on each CDF. A zoomed range-image (bottom) shows each raypath and its surrounding neighborhood.*

In addition to recording range CDFs, Figure 6 also includes triangle markers plotting measured reflectance (right-hand vertical axis) as a function of measured range (horizontal axis). A reflectance value of 100 describes a perfect Lambertian reflector (Velodyne, 2019) viewed normal to the surface, where the surface fills the beam cross section. Values below 100 indicate imperfect Lambertian surfaces, with the linear scaling between 0 and 100 corresponding to the percentage power scattered back. A returned power percentage below 100 may indicate absorption, scattering from an oblique surface, or scattering from a surface that does not fill the beam cross-section. Values above 100 are possible, too, indicating strong specular reflections, generally from a retro-reflective target.

The first three regions of interest (ROI) in the Volpe data look Southeast from the van, toward the edge of the parking lot. ROI 1 centers on a hedgerow, where the foliage is dense enough that the CDF resembles a step, similar to the *wall* case from the Newer College dataset. The 50 sequential range samples for the central raypath of ROI 1 have a standard deviation of 1.3 cm. ROI 2 and ROI 3 focus on the middle and edge of a tree. The CDFs for these cases indicate multiple scatters, similar to the CDF for the *foliage* case from the Newer College dataset. For ROI 2, there is a notable difference in estimated reflectance between the nearer returns (below 13.3m) and farther returns (above 13.4 m). The nearer returns show reflectance levels as high as 20%, whereas the farthest returns are below 5%. It is reasonable to infer that the higher reflectance values are caused by the beam striking an object with a larger cross section (e.g. the branch of the tree) and the lower values, an object with smaller cross section (e.g. a leaf). For ROI 3, at the edge of the tree where no large branches are expected, all reflectance values are small (between 2% and 4%).

The next three raypaths in the Volpe data, ROI 4 through ROI 6, North of the data-collection van into the parking lot. ROI 4 and ROI 6 both strike a tree and their CDFs observe multiple returns, separated by as much as 2 m for ROI 4 and 1.5 m for ROI 6. By contrast, ROI 5 describes a retroreflective sign (reflectance values between 179 and 183) with a step-like CDF, similar to the wall case from the Newer College data. The range values of ROI 5 have a standard deviation of only 0.8 cm.

The data show a slight distinction between raypaths through the middle and edges of a tree. At the center of a tree (like ROI 2), ROI 4 observes one cluster (possibly a branch) with reflectance as high as 30% and other clusters (possibly leaves) with reflectance below 10%. At the edge of a tree (like ROI 3), ROI 6 observes only low reflectance values (below 10%), which might be expected for the edge of a tree where branch tips are small.

### 3.2  Comparison of Temporal and Spatial CDFs

At a frame rate of 10 Hz, resolving a temporal CDF along a given raypath requires a second or more. By comparison, a spatial CDF can be computed instantaneously for a local neighborhood. Given the ease of constructing a spatial CDF in real time, it is reasonable to ask if the spatial CDF centered on a raypath is predictive of the raypath's temporal CDF.

A CDF comparison for the Newer College data is shown below in Figure 8. Spatial and temporal CDFs are plotted for four raypaths: wall, corner, window, and foliage. The temporal CDF is shown in blue, and the spatial CDF is shown in orange. The spatial CDF is generated from the first frame of the lidar sequence, compiling range values from a 5x5 neighborhood around the raypath of interest. Note that the x-scale is non uniform between raypaths, to maximize the zoom level for each case.

The spatial CDFs exhibit remarkable similarity to their temporal CDFs for these four representative examples from the Newer College dataset. The comparison is particularly close for the wall example, where the raypath is nearly normal to the plane of a large, solid wall. The comparison between spatial and temporal CDFs is less precise but still reasonably close for the corner and foliage raypaths. The CDFs for the window case are also similar, in the sense that both the spatial and temporal CDF observe three distinct clusters of points; however, the CDFs diverge in the frequency of occurrence of each cluster. Notably more points scattered from the window surface (near 11 m) for spatial CDF than the temporal CDF.

A comparison of temporal and spatial CDFs for the Volpe data is shown in Figure 9. CDFs are shown for the six raypaths of interest, labeled ROI 1 through ROI6. For the Volpe dataset, spatial CDFs were computed over a 3x3 neighborhood around the raypath of interest (see Figure 4). Also, the spatial CDFs in Figure 9 were computed not just for the first frame of the sequence, but for all 50 frames of the sequence. The data points from all 50 spatial CDFs (orange dots) are superimposed on the temporal CDF (blue).

The spatial CDFs for a given ROI tend to identify the same clusters through time, but not always with the same frequency. For ROI 2, for example, three clusters of points are consistently visible in the spatial CDFs (with clusters centered at 13.3, 13.6, and 13.8 m in range); however, the number of points distributed across each of the three clusters was variable from one spatial CDF to the next.

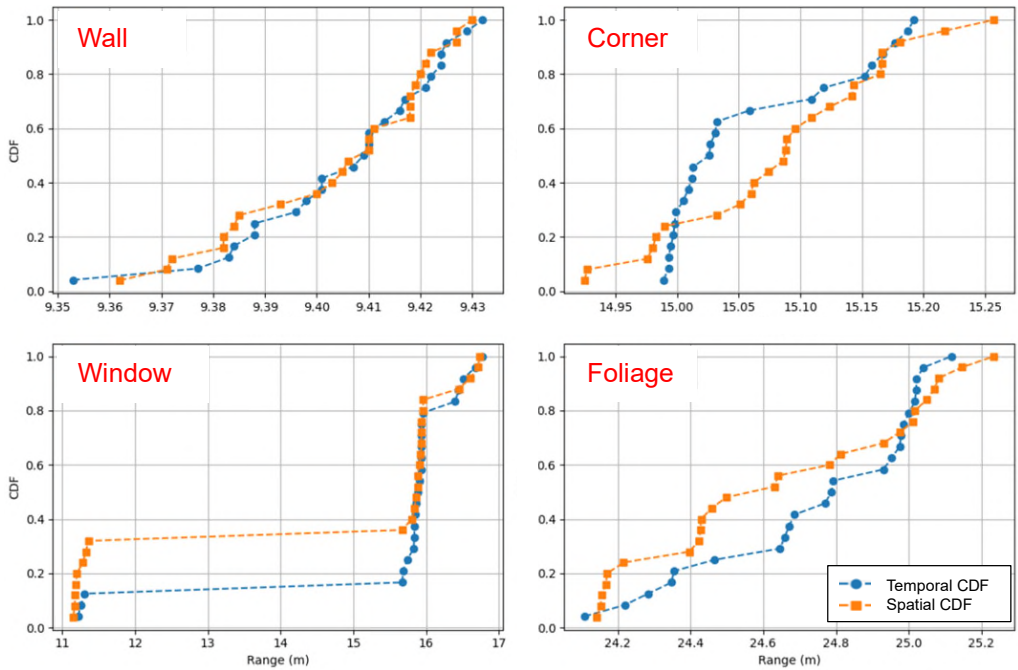

**FIGURE 8**
*Spatial CDFs (orange) are compared to temporal CDFs (blue) for the four raypaths analyzed from the Newer College Dataset.*

Spatial and temporal CDFs compare favorably for the Volpe data, but less well than for the Newer College data. For the tree scenes visualized in the Volpe data (ROI 2, ROI 3, ROI 4 and ROI 6), the thin leaf coverage consistently resulted in the spatial CDF identifying additional clusters not identified by the temporal CDF. In all of these foliage cases, the closest ranges identified were the same for both the spatial and temporal histograms. However, the spatial histograms also captured intermediate clusters and a more distant clusters, one falling slightly beyond the farthest range measurements of the temporal CDF. These "outlier clusters" appeared for the spatial CDFs for all four tree ROIs. Generally, the outlier clusters from the spatial CDF fall within 0.5 m of the most distant cluster in the temporal CDF, but ROI 2 has an outlier cluster at 16 m, nearly 3 m beyond the most distant cluster of the temporal CDF. From visual context, this 3 m gap appears to separate the foreground tree from a second tree behind it.

Another interesting CDF comparison from the Volpe data involves ROI 1, which captures a hedgerow to the side or the parking lot. Though the top of the hedge is rounded, as shown in Figure 3(d), the top surface can reasonably be modeled as a flat plane which the central raypath approaches at a highly oblique incidence (and elevation) angle of -9°. As compared to a central raypath at elevation -9°, the vertical pixel neighbors have elevations of -7° and -11°. At a range of about 21 m, this 2° difference in the oblique raypath to a horizontal plane changes the measured range by about 0.7 m. This model appears to describe the spatial CDF for ROI 1, where range measurements are blurred over a span of ranges that fall about 0.5 m on either side of the central raypath (which was characterized as step-like for the temporal CDF).

Of the ROIs considered for the Volpe dataset, the most consistent is ROI 5, associated with a road sign. This object was a relatively flat, with its surface oriented close to perpendicular to the raypath; for a flat object normal to the raypath, we expect the spatial and temporal CDFs will be well matched, a result borne out by the ROI 5 data in Figure 9.

### 3.3 Effect of Motion Compensation on Temporal CDF

Though we did not run extensive analysis using the motion compensation strategy described by (6), we did apply the methodology to one case (the window case from the Newer College dataset) to observe whether motion compensation would change the character of the CDF. Uncompensated and motion-compensated versions of the window case CDF are shown in Figure 10.

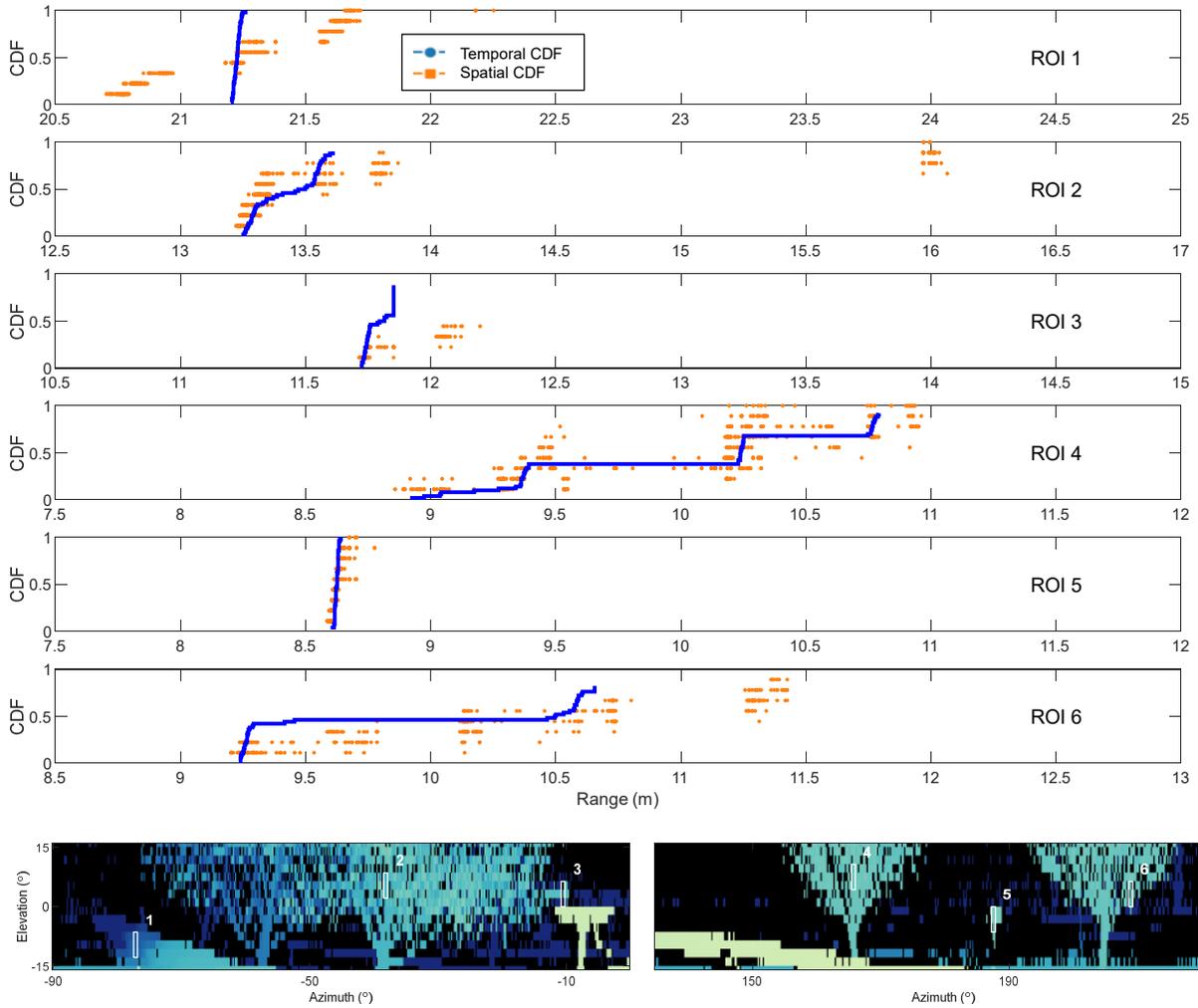

**FIGURE 9**
*50 sequential spatial CDFs (orange dots) are compared to temporal CDFs (blue) for the six raypaths from the Volpe Dataset.*

The primary effect of the motion compensation in this case appears to be a blurring effect, which spreads out the boundaries of the distinct clusters observed in the uncompensated case. It is important to note that the motion compensation relied on an estimate based purely on a lidar-image correspondence in the region of the raypath. It is not yet clear whether this form of motion compensation provides a more accurate representation of the temporal CDF or not. Future efforts will be needed to further explore the possible benefits of motion compensation, by using a separate sensor (for example, an inertial navigation system) to obtain the motion correction.

For the window case, an interesting observation is that the motion-compensation provides evidence that the window glass produced some returns. Since a lattice is embedded in the window, it could be possible that several returns from the window surface are generated by the lattice rather than the transparent window glass. (See lidar image of window in Figure 5). However, the motion compensation consistently moved the raypath of interest on to transparent glass and away from the lattice. This detail is relevant because three shorter-range measurements (near 11 m) were observed on the window surface, both before and after motion compensation. The fact that window surface measurements persist, even after motion compensation to steer the samples toward window glass, suggests the windows are not fully transparent to the lidar beam.

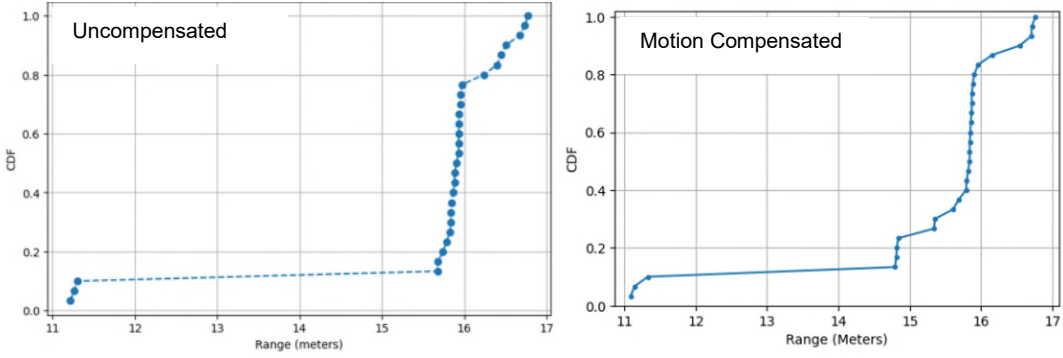

**FIGURE 10**
*Temporal CDF for window raypath from the Newer College Dataset: the CDF produced without motion compensation (left) is compared to the CDF with motion compensation (right).*

## 4 DISCUSSION

This section considers how the results shown in the prior section might impact lidar-based automotive localization. Specifically, we consider (i) a conceptual model for the range-error distribution for a multi-return raypath, (ii) the impact of multi-return raypaths on lidar registration algorithms, and (iii) methods for detecting multi-return raypaths.

### 4.1 Distribution Modeling

For analysis of lidar-based positioning, it is helpful to develop a conceptual model for the range measurement in the case of a multi-return raypath. The key detail is that the signal returning to the detector may have scattered from any of a set of surfaces contained within the diverging cross-section of the lidar beam (see Figure 1). Even a single-return raypath has measurement noise (e.g. 2 cm standard deviation random noise for the wall case of the Newer College data); however, we expect the multi-return raypath to exhibit random noise within multiple clusters, one for each scatterer along the raypath.

As a first-order conceptual model, we can consider each cluster of points to be represented by a Gaussian probability density function (PDF). By extension, the overall range-measurement PDF for the multi-return raypath might reasonably be modeled as a Gaussian Mixture Model (GMM). We can write an expression for the range-distribution density function $f$ by assuming a GMM with a distinct weight $\alpha_l$, a mean $\mu_l$, and a standard deviation $\sigma_l$ for each cluster $l$:

$$f(x) = \sum_{l=0}^{L-1} \alpha_l \, \mathrm{N}(x;\mu_l,\sigma_l) \,. \tag{7}$$

The GMM is expressed here as a summation of Gaussian (or *normal*) distributions N for each cluster, with $L$ being the total number of clusters. To ensure that the PDF model $f$ integrates to one, the weights are constrained to sum to one, with

$$\sum_{l=0}^{L-1} \alpha_l = 1. \tag{8}$$

The integral of the PDF is the CDF. For a GMM, the CDF model $F$ is represented as a sum of normal-distribution CDFs, for which we use the symbol $\Phi$:

$$F(x) = \sum_{l=0}^{L-1} \alpha_l \, \Phi(x;\mu_l,\sigma_l) \,. \tag{9}$$

It is somewhat challenging to use this model predictively for a given raypath; however, it is not hard to use the model descriptively for an existing empirical CDF. Consider the case of ROI 4 from the Volpe dataset (see Figure 7). A GMM model for the temporal CDF can be generated easily, as shown in Figure 11. Because the clusters are relatively distinct, a reasonable GMM approximation can be obtained by segregating the data at CDF-value thresholds ( at 0.14, 0.38, and 0.68 for this case). Once the data are segregated, a weight, mean, and standard deviation can be trivially computed for each cluster. The resulting model PDF has four clusters, shown in green at the top of Figure 11. The bottom of the figure shows that the CDF model (also in green) closely tracks the empirical CDF (blue dots). Though not used here, more sophisticated methods like Expectation Maximization have been defined to infer GMM models for overlapping clusters (Yao & Xiang, 2024).

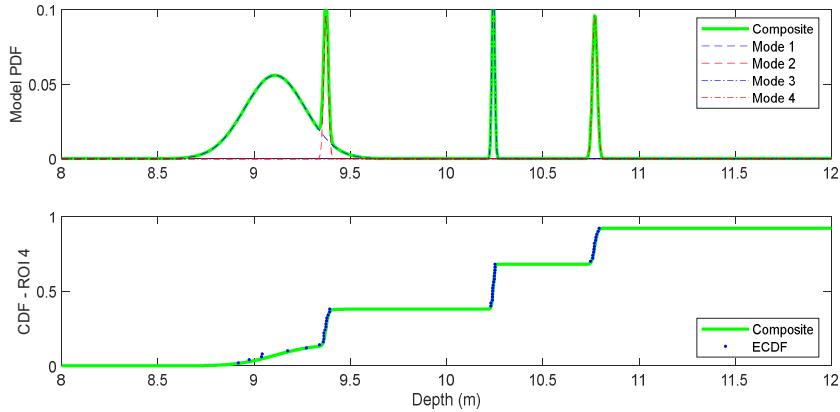

**FIGURE 11**
*A Gaussian Mixture Model for ROI 4 from the Volpe Dataset: a PDF model is generated by data segmentation (top) and the associated CDF model compared well to the empirical CDF (bottom)*

**4.2   Impact of Multi-Return Raypaths on Lidar Scan and Map Matching**
For localization applications, an impactful issue is the degree to which probabilistic, multi-return cases impact algorithm accuracy. In this section, we conduct a qualitative analysis of the effects of probabilistic returns on scan and map matching algorithms. For visualization purposes, we consider two classes of algorithm: first, feature or point matching algorithms including variants of the Iterative Closest Point (ICP) algorithm of Besl & McKay (1992) and, second, voxel-based algorithms including variants of the Normal Distribution Transform (NDT) algorithm of Biber & Straßer (2003). As a starting point for qualitative discussion, in this section we consider only multi-return CDFs consisting of two clusters of points separated by a gap $\Delta$, somewhat like the window case shown in Figure 6. The effects of probabilistic returns depend not only on the registration algorithm (e.g. ICP or NDT) but also on the gap distance $\Delta$ and the type of reference image used (i.e., prior scan or a map).

Let's begin by considering the case when the CDF gap is large, with a separation of several meters between clusters (as in the window case of Figure 6). If, for a particular raypath, the current scan contains a point from one cluster (e.g. the window surface) while the reference contains only a point from the other cluster (e.g. the room interior), then the points from the current scan and reference will not be associated during the registration process. Instead, this raypath will be mis-associated with other points in the point cloud. This mis-association is illustrated in Figure 12(a) for ICP and in Figure 13(a) for NDT. In the figures, probabilistic returns are designated by two cluster locations (dashed circles) along the same raypath. A measurement along that raypath could appear in either circle, probabilistically. Assume that both current and reference scan returns (triangle and cross markers) exist along the raypath but at different cluster locations. Although ICP, NDT, and other registration methods attempt to align point clouds, so that related returns are as close as possible, the large gap prevents registration between the current and reference samples for the same raypath. ICP, for example, operates by explicitly connecting each current scan point to the nearest reference scan point and then adjusts scan registration to minimize the distance between all matched pairs. Though points acquired along the same raypath would normally be associated by ICP, the large gap causes the two points to instead be associated randomly with a nearby point in the cloud. NDT, as another example, uses a voxel-based registration strategy, which attempts to align the centroid of the current points in a voxel with the centroid of the reference points in the same voxel. When the current and reference scan measurements along a raypath are separated by a large gap, these measurements fall into distinct voxels, as shown in Figure 13(a). The result is that each measurement impacts the statistics for its assigned voxel (e.g. each shifting the computed centroid slightly). For both the NDT and ICP cases, it seems reasonable to model these large-gap cases as introducing a uniformly distributed error over a region of association (e.g. over the maximum matching radius for ICP-like algorithms or over voxel radius for NDT-like algorithms). This uniform model reflects the notion that the random association with a neighbor or with other points in a voxel will pull the registration by a bounded distance in a random direction.

The outcome is slightly different for a two-return raypath with a small gap between the scattering surfaces. For these cases, the current and reference scan samples from the raypath might be displaced, as shown in Figure 12(b) and Figure 13(b), but not displaced enough to prevent association (in ICP) or placement in the same voxel (in NDT). The small gap creates an additional measurement error along the raypath direction. In essence, this case is just a conventional ranging error, but with a bias (gap size) added to the conventional random noise.

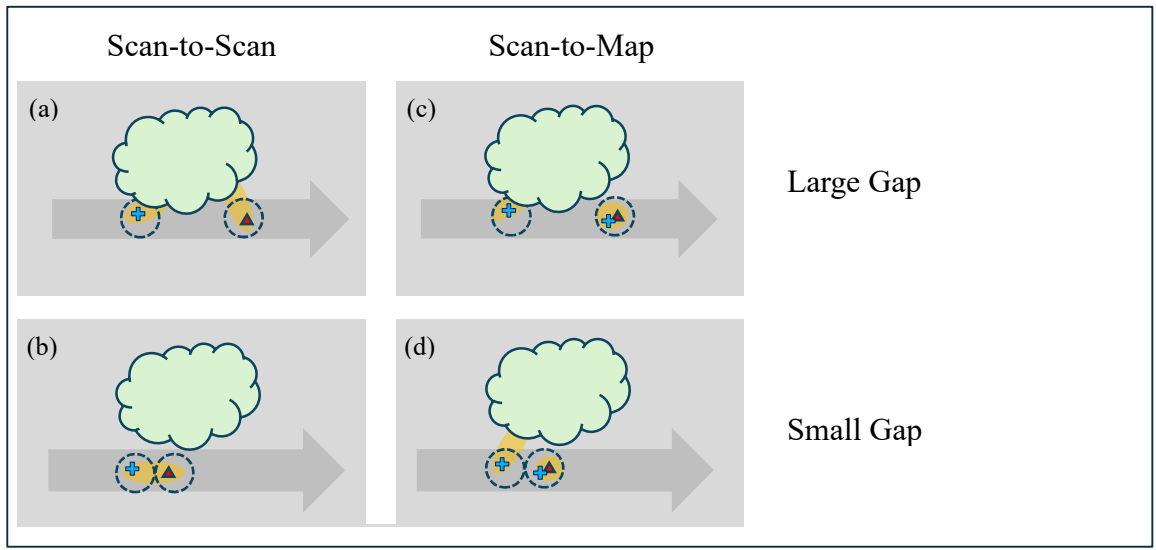

**FIGURE 12**
*Qualitative impact of probabilistic returns on scan and map matching for feature-matching algorithms like ICP*

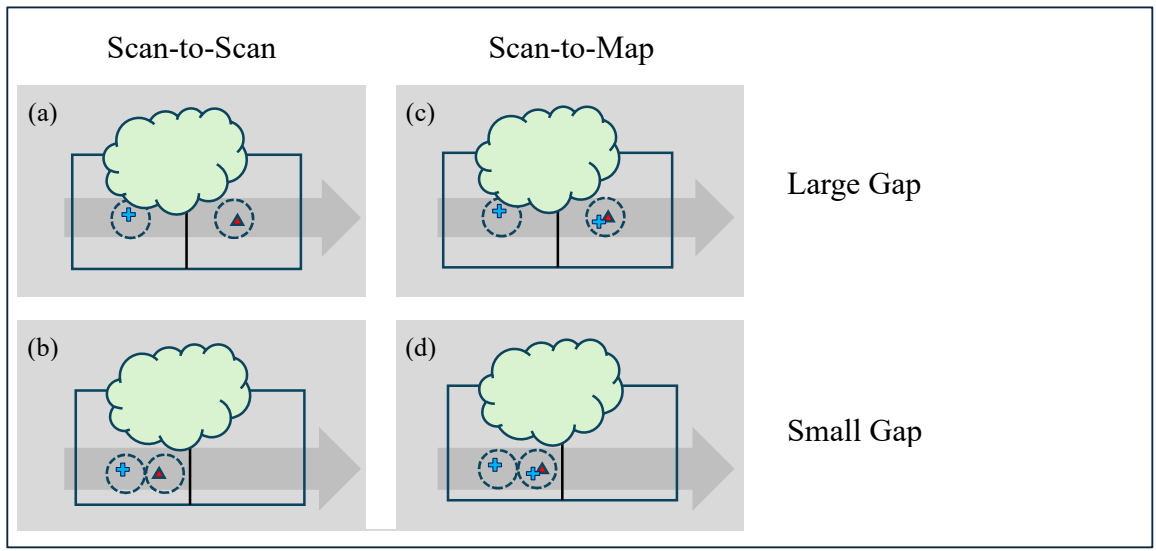

**FIGURE 13**
*Qualitative impact of probabilistic returns on scan and map matching for voxel-based algorithms like NDT*

Whether the current scan is matched to a reference scan or a georeferenced map, the error analysis is largely the same, with one exception. The exception is that a map (for example, a high-definition or HD map) can contain more than one return for a given raypath. To model this effect qualitatively, consider the case when the reference map includes a point (cross marker) associated with each cluster (dashed circle) along a ray path, as represented in the right column of Figure 12 and Figure 13. In this case, the current-scan measurement (triangle marker) will always fall near one of the map points. In the cluster (dashed circle) where the map and current-scan points appear together, they will likely be associated (ICP) or fall in the same voxel (NDT), keeping the associated registration error at its nominal value. However, the extra (unmatched) reference point, could potentially generate an incorrect match (ICP) or bias the statistics in its voxel (NDT). In the worst case, this extra point could introduce an error similar to the large gap case of Figure 12(a) and Figure 13(a), so it would be conservative to use a similar uniform-distribution model for the scan-to-map cases of the right column of Figure 12 and Figure 13. In the future, a tighter (less conservative) model might also be considered. Also, additional analysis should be considered for explicitly probabilistic maps, maps which inherently represent each raypath as a probabilistic CDF (McDermott, 2025).

### 4.3 Toward a Multi-Return Monitor using Spatial CDFs

If multi-return raypaths can introduce elevated errors for localization, as described in the prior section, then it is reasonable to exclude some or all multi-return cases if they can be detected. Though collecting data for a temporal CDFs is prohibitive (3 s lag for our Newer College analysis and 5 s lag for our Volpe analysis), there is no latency associated with computing a spatial CDF. To the extent that spatial CDFs are a reasonable predictor for temporal CDFs (see Figure 8 and Figure 9) a monitor could potentially be constructed to exclude raypaths when the associated spatial CDF exhibits high variability. Future work will be needed to fully explore whether such a monitor is both feasible and useful.

## 5 SUMMARY

In this paper, we applied lidar data to characterize raypaths which probabilistically report multiple returns. We observe, by analyzing mechanically rotating lidar data, several types of raypath that generate multiple returns, including for porous objects (foliage) and transparent objects (windows). We observe that the empirical cumulative distribution functions (CDFs) of multi-return raypaths consist of multiple clusters, each centered at different range. In a representative example, these clusters were reasonably well described (analytically if not predictively) by a Gaussian mixture model. Building on this mixture model concept, we perform a qualitative analysis of scan and map-matching error for multi-return raypaths. We also proposed a possible strategy to monitor for multi-return raypaths by using a spatial CDF.

## ACKNOWLEDGEMENTS

The authors wish to acknowledge and thank the U.S. Department of Transportation Volpe Center and PNT & Spectrum Management Office for sponsorship of this work. We also gratefully acknowledge KBR and Tufts University, which supported specific aspects of this research. Opinions discussed here are those of the authors and do not necessarily represent those of the USDOT, KBR, or other affiliated agencies.